\documentclass[a4paper]{spie}
\usepackage[]{graphicx}
\usepackage{mathtools}
\usepackage{url}
\usepackage{multicol}
\usepackage{caption}
\usepackage{subcaption}
\usepackage[polish,english]{babel}

\title{Cataract influence on iris recognition performance} 

\author{Mateusz Trokielewicz\supit{1,2}, Adam Czajka\supit{2,1}, Piotr Maciejewicz\supit{3}
\skiplinehalf
\supit{1}Biometrics Laboratory, Research and Academic Computer Network,\\ Wawozowa 18, 02-796 Warsaw, Poland;\\
\supit{2}Institute of Control and Computation Engineering, Warsaw University of Technology, Nowowiejska 15/19, 00-665 Warsaw, Poland;\\
\supit{3}Department of Ophthalmology, Medical University of Warsaw,\\ Lindleya 4, 02-005 Warsaw, Poland 
}

\authorinfo{Correspondence: mateusz.trokielewicz@nask.pl, aczajka@elka.pw.edu.pl}
 
\begin{document} 
  \maketitle 

\begin{abstract}
This paper presents the experimental study revealing weaker performance of the automatic iris recognition methods for cataract-affected eyes when compared to healthy eyes. There is little research on the topic, mostly incorporating scarce databases that are often deficient in images representing more than one illness. We built our own database, acquiring 1288 eye images of 37 patients of the Medical University of Warsaw. Those images represent several common ocular diseases, such as cataract, along with less ordinary conditions, such as iris pattern alterations derived from illness or eye trauma. Images were captured in near-infrared light (used in biometrics) and for selected cases also in visible light (used in ophthalmological diagnosis). Since cataract is a disorder that is most populated by samples in the database, in this paper we focus solely on this illness. To assess the extent of the performance deterioration we use three iris recognition methodologies (commercial and academic solutions) to calculate genuine match scores for healthy eyes and those influenced by cataract. Results show a significant degradation in iris recognition reliability manifesting by worsening the genuine scores in all three matchers used in this study (12\% of genuine score increase for an academic matcher, up to 175\% of genuine score increase obtained for an example commercial matcher). This increase in genuine scores affected the final false non-match rate in two matchers. To our best knowledge this is the only study of such kind that employs more than one iris matcher, and analyzes the iris image segmentation as a potential source of decreased reliability. \let\thefootnote\relax\footnote{Accepted for publication in the Proceedings Volume 9290, Photonics Applications in Astronomy, Communications, Industry, and High-Energy Physics Experiments 2014; 929020 (2014) https://doi.org/10.1117/12.2076040
Event: Symposium on Photonics Applications in Astronomy, Communications, Industry and High-Energy Physics Experiments, 2014, Warsaw, Poland}
\end{abstract}

\section{Introduction}
The assertion that the human iris pattern is resilient and does not undergo changes throughout one's lifetime has remained very strong in biometric community. Safir and Flom mention this for the first time in their 1987 patent that paved the way for iris recognition: 'significant features of the iris remain extremely stable and do not change over a period of many years' \cite{SafirFlom}. John Daugman supports this in his patent claim from 1994, stating that 'the iris of the eye is used as an optical fingerprint, having a highly detailed pattern that is unique for each individual and stable over many years', as well as 'the iris of every human eye has a unique texture of high complexity, which proves to be essentially immutable over a person's life' \cite{DaugmanPatent}. 

Temporal stability of the \emph{pattern} present in a healthy iris does not necessarily mean that it cannot undergo changes resulting from external factors. Cited claims assume that the eye has not been subject to disease, injury or other trauma that may result in altering the iris pattern or obstructing its appearance, and thus making the recognition not feasible. Although Daugman acknowledges that 'as an internal organ of the eye, the iris is well protected from the external environment' \cite{DaugmanPatent}, there may be trauma situations in which this protection is not sufficient, or an eye-affecting illness takes place. Safir and Flom were aware of this issue and expressed it in their patent claim: 'A sudden or rapid change in such a feature may result in a failure to identify an individual, but this may alert the individual to the possibility of pathology of the eye' \cite{SafirFlom}.

The ISO/IEC 29794-6 standard specifies medical conditions that may affect iris or eye structures surrounding it in a way that will cause problems with iris recognition systems \cite{ISO}. These include excessive dilating or constricting the pupil (induced by disease, trauma, drugs or alcohol), diseases affecting the iris and/or the cornea (\emph{e.g.} iritis, microcornea, megalocornea, microbial keratitis, leukoma), congenital diseases (partial or total aniridia), surgical procedures (cataract surgery, iridotomy and iridectomy) and many other pathologies (bleedings, arcus senilis, Kayser-Fleischer ring). The aforementioned document also specifies two scenarios of how ocular diseases can influence the iris recognition:

\begin{itemize}
\item injury or disease happen to an already enrolled biometric system user, and therefore may contribute to the degradation of similarity scores between images obtained afterwards with present illness and gallery images without such changes (\emph{e.g.} a person undergoes an iris-affecting eye surgery, such as iridectomy); in severe cases this may result in false rejecting the user by the system; 
\item a disease or injury occurs before the first enrollment (\emph{e.g.} a congenital disease, such as complete lack of iris -- \emph{aniridia}, is present); this may degrade the overall performance of an iris recognition system or even render it unusable for this particular person.
\end{itemize}

In this paper we focus on the latter case. We aim at answering the question whether an iris recognition system may degrade its performance when presented with images obtained from patients suffering from cataract. 

\section{Medical prerequisites}

\subsection{Cataract characteristics}

\begin{figure}
\centering
\begin{subfigure}{0.5\textwidth}
  \centering
  \includegraphics[width=0.99\linewidth]{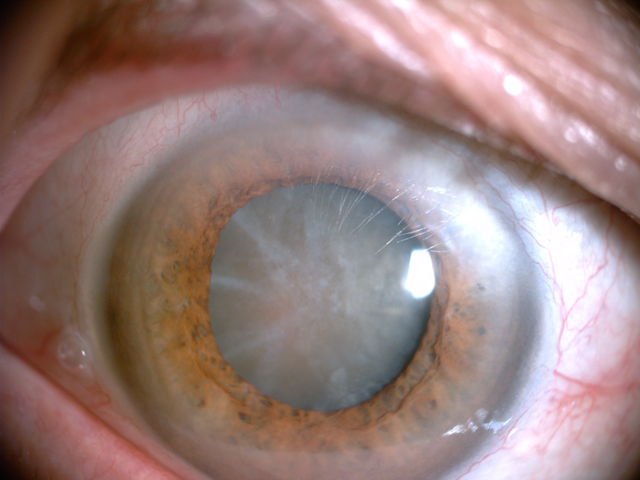}
\end{subfigure}%
\begin{subfigure}{0.5\textwidth}
  \centering
  \includegraphics[width=0.99\linewidth]{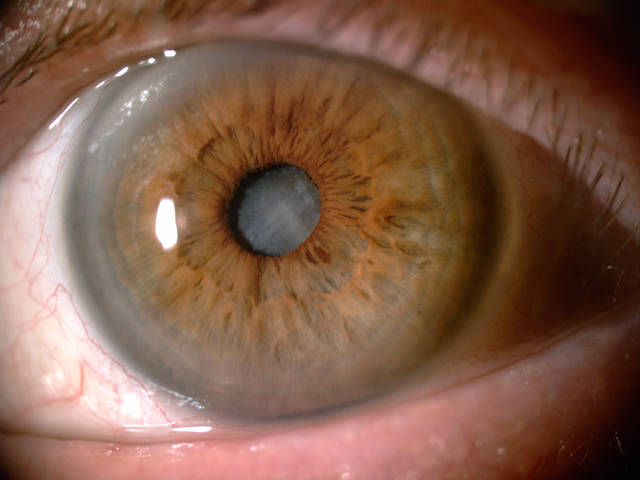}
\end{subfigure}
\vskip0.2cm
\caption{Cataract-affected eyes. Clouded lens prevents the light from entering the eyeball. In addition, age-related, ring-shaped corneal opacity (\emph{arcus senilis}) is present.}
\label{cataract}
\end{figure}

Cataract is a clouding of the eye lens that results in worsening the eyesight. The most common symptoms are cloudy or blurry vision, faded colors, poor night vision and multiple images in one eye. Visual loss occurs due to opacity of the lens that obstructs the light from passing through and being focused onto the retina, Fig. \ref{cataract}. It can occur in either one or both eyes. Cataract is a significant global problem since it is the leading cause of visual impairment and accounts for 30\% of blindness worldwide, mostly in developing countries.

Cataract is a multifactorial disease associated with age, genetic predisposition, smoking, diabetes mellitus, diet, drug intake or exposure to ultraviolet-B radiation. The combination of interactions and subliminal exposures to many factors over a person's lifetime can contribute to the development of the most common, age-related cataract. The oxidative stress induced by light penetrating the intraocular space and consequent photochemical reactions inside the lens are believed to be the main cause. Secondary cataract is much less common and can occur after the eye injury, intraocular inflammation or exposure to ionizing radiation. Congenital cataract occurs in babies born with cataracts or develops in early childhood, frequently in both eyes.

The only effective treatment to prevent blindness induced by lens opacity is a surgical cataract extraction followed by implantation of an intraocular artificial lens. Other methods such as medications or diet have not been shown to stop cataract progression. During cataract surgery called \emph{phacoemulsification}, a small incision is made on the side of the cornea, then a tiny probe inserted into the eye emits ultrasound waves that crush and break the lens material so that it can be removed using suction. After the anatomic lens has been removed, it is replaced by an artificial intraocular lens with individually calculated focusing power.

\subsection{Influence on remaining eye structures}

\begin{figure}
\centering
\begin{subfigure}{0.5\textwidth}
  \centering
  \includegraphics[width=0.99\linewidth]{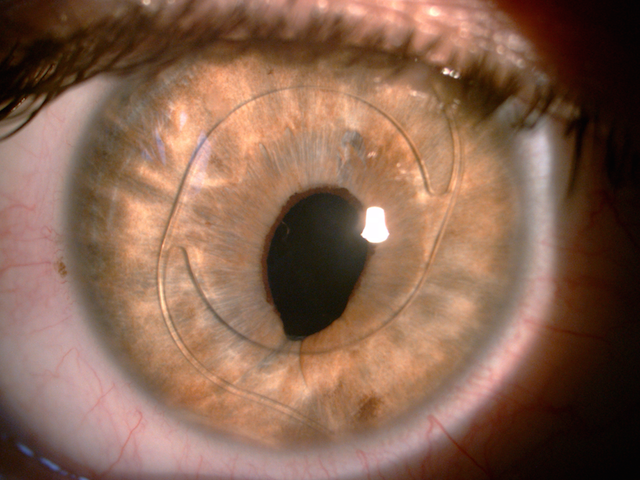}
  \caption[outside_lens]
  { \label{outside_lens} 
  Lens implant placed in front of the iris.}
\end{subfigure}%
\begin{subfigure}{0.5\textwidth}
  \centering
  \includegraphics[width=0.99\linewidth]{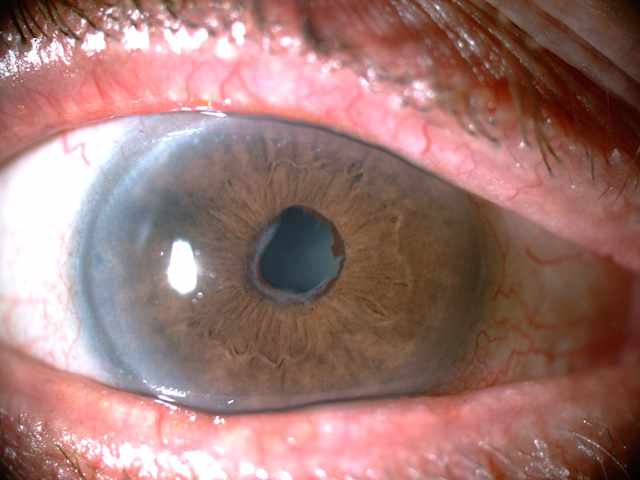}
  \caption[anterior_synechiae]
  {\label{anterior_synechiae}
  Posterior synechiae.}
\end{subfigure}
\vskip0.2cm
\caption{When the lens implant is placed behind the cornea and in front of the iris, it may distort the way the iris looks and also the shape of the pupil (a). Anterior or posterior synechiae may accompany cataract and cause a deviation from the pupil's circularity (b).}
\label{cataract_accompanying}
\end{figure}

The perfect condition following cataract extraction is that the intraocular lens is located behind the iris, intracapsulary in posterior chamber of the eye. Usually it does not influence the shape of the pupil and the surface of the iris. Following complicated cataract surgery there may be insufficient remaining capsular support for intracapsular placement of the artificial lens. In such cases the lens can be located in front of the iris, in anterior chamber of the eye using sulcus placement or iris fixation. In this method the lens is attached to mid-peripheral part of iris with claw shaped hooks. An intraocular implant of this type can distort the shape of the pupil and alter look of an iris, Fig. \ref{outside_lens}.

There are many other conditions accompanying different types of cataract, potentially affecting iris surface, \emph{e.g.}: acute glaucoma, anterior and posterior synechiae (as shown in Fig. \ref{anterior_synechiae}), rubeosis iridis (pathological vascularization in the iris), iris atrophy, pseudoexfoliation syndrome, conditions after iris laserotherapy or post traumatic cataract.

\section{Related work}
Because of difficulties in creating biometric databases epitomizing illnesses, surgical procedures and eye injuries that may happen to a biometric system user, there is very limited research on the subject. These investigations mostly cover small datasets and few illnesses (usually one).

T. M. Aslam \emph{et al.} conduct a study concerning ocular disease influence on iris recognition using data collected from 54 patients with various eye conditions \cite{Aslam}. Dataset consisted of images of 69 eyes with illness present and a control group of 39 healthy eyes. Patients had their eyes photographed using IrisGuard H100 iris recognition camera during their first visit, then a surgery or other form of treatment was performed and the eye was photographed once again afterwards. Daugman's method-based algorithm was used to handle the iris encoding and Hamming distances were calculated between templates acquired before and after the treatment. The following ocular conditions were included: glaucoma treated with laser iridotomy, anterior uveitis (inflammation of the iris), infective and non-infective corneal pathologies, episcleritis, scleritis and conjunctivitis. 

Experiments showed that the evaluated method was resilient to most conditions except for the iris-affecting anterior uveitis (iritis). 5 out of 24 irises in this subset were false rejected, yielding a false non-match rate (abbreviated FNMR later on) of about 21\% with acceptance threshold set to a Hamming distance value of 0.33. However, authors indicate that in each of the cases of incorrect rejection, pupil-dilating pharmaceutics were applied. In addition, two patients had visible changes in the cornea and large anterior chamber activity, while the other three suffered from posterior synechiae interrupting circularity of the inner iris boundary.

Researchers also present statistically significant increase in average Hamming distances derived from anterior uveitis affected subset when compared to the control group ($p$-value $< 10^{-4}$ for the null hypothesis that the average Hamming distances in those two groups are equal). At the same time, there were no statistically significant differences between mean match scores for other pathologies. Aslam also suggests that lack of perceivable impact of corneal conditions on match score might be derived from different transmission of near-infrared light (abbreviated NIR later on) through eye structures. It is less attenuated by disease-induced  objects present inside cornea and anterior chamber of the eye, therefore allowing the underlying structures of the iris to be successfully imaged. As for the lack of laser iridotomy impact, it can be explained by the fact that the puncture made by laser radiation is small enough not to alter the iris pattern in a way that would cause recognition errors. 

As a possible cause of recognition failures for selected cases authors propose an explanation referencing to a combination of inner iris boundary's deviation from circularity due to posterior synechiae and excessive, pharmacologically induced pupil dilation. It is suggested that the \emph{rubber sheet model} proposed by Daugman \cite{DaugmanArticle} that assumes linear, elastic change of iris pattern with changing pupil diameter is adequate enough when those changes are physiological, but insufficient when pupil dilation is pharmacologically induced. Other possible cause of degraded method’s performance might be affiliated with changes in the iris itself that are not visible during slit lamp examination, but reveal themselves when near infrared illumination is applied. In addition, posterior synechiae can disturb pupil circularity and thus create issues with image segmentation.

Roizenblatt \emph{et al.} \cite{Roizenblatt} describe an experiment in which 55 cataract patients were enrolled to the LG IrisAccess 2000 biometric system before undergoing a cataract surgery, and then authenticated 30 days after the treatment to make sure that all healing processes came to an end, and additionally 7 days after mydriatics (pupil-dilating drugs) were no longer administered. After 30 days differences in pupil diameter when compared to the images acquired before the procedure were no larger than 1.5 mm. In addition to that, each photograph has been evaluated by an ophthalmology specialist and given a score between 0 and 4, depending of the visual change present in the iris. One point was given for each of the following: depigmentation, pupil ovalization, focal atrophy with transillumination, focal atrophy without transillumination.

Researchers report a noticeable correlation between visible iris pattern alteration assessed by an ophthalmologist and change in Hamming distance between iris codes derived from pre- and post-surgery images. In six cases images were false rejected. Although other iris were classified correctly, there were significant differences in Hamming distances and visual evaluation scores. Authors describe change in average HD from 0.098 to 0.2094 (hence an increase by 11.3\%) and average visual score growth of 11.13\%. There were also statistically significant differences in HD between groups of eyes with highly varying visual scores.

Atrophy-related changes that take place during the cataract procedure are proposed as a possible cause of degradation in recognition performance. It is known that energy deployed in the eye during the surgery may lead to depigmentation and atrophy of the iris tissue, but underlying physiological mechanism is not fully studied. To avoid cataract surgery related errors in biometric systems, a re-enrollment is suggested when slit lamp examination reveals visible changes in the iris pattern.

In an experiment conducted by Dhir \emph{et al.} \cite{Dhir} in addition to cataract surgery itself, an influence of mydriatics is studied. 15 patients have been enrolled in the system and then a verification was performed 5, 10 and 15 minutes after mydriatics were applied and also 2 weeks after the procedure. 100\% accuracy was achieved when using images obtained after the surgery, however, it is crucial to acknowledge that authors do not include images with pre-existent corneal and iris pathologies, or those that represent iris damaged during the procedure, in the final dataset. They suggest that problems with iris recognition after the surgery may originate from slight displacement of the iris towards the center of the eyeball that is caused by installing a thinner artificial lens implant, as well as from increased number of specular reflections effecting from the implant (what may lead to trouble using certain localization methods).

When it came to the images acquired after the use of mydriatics, the recognition performance degraded as the pupil diameter increased, leading to false rejection in 6 out of 45 cases (hence FNMR=13.3\%). Authors found Hamming distance to systematically grow as the time after the application of mydriatics elapsed. Noticeably, this gives lawbreakers an opportunity to use this phenomenon as a method for deceiving an iris recognition system by using such pharmaceuticals to register under multiple identities, hence the system should make an alarm when observing unusually dilated pupils.

Separate group of medical procedures involving the eye are laser-assisted refractive defect correction, that are used widely across the globe to minimize discomfort in patients suffering from myopia, hypermetropia and astigmatism. This involves cutting the cornea to create a flap with a hinge left on one side of it. The flap is lifted to expose the middle part of the cornea, which is then ablated using pulses from 193 nm excimer laser to achieve a finely tuned shape that depends on the treated condition. The corneal flap is then folded back and eye is left to heal itself (\emph{Laser-Assisted in-Situ Keratomileusis} method, abbreviated \emph{LASIK}). Yuan \emph{et al.} \cite{Yuan} carry out a study to find out if those changes are able to alter the iris pattern and therefore result in affecting performance of an iris recognition system. This study incorporated a group of 7 people that were to be subject to the LASIK procedure on both eyes (14 eyes total). All irises have been imaged using a camera developed by authors, and encoded using Masek's algorithm. Out of those 14 irises, 13 were recognized correctly after the procedure. However, the one iris that failed to be matched afterward had significant differences in pupil diameter and a visible deviation from pupil circularity. Authors argue that these aspects may be associated with healing processes that occur in the eye after the procedure is performed. LASIK surgeries are said to have minimum influence on the tested biometric system, however further studies including larger datasets are called for.

Borgen \emph{et al.} \cite{Borgen} apply a different approach in analyzing ocular disease impact on iris recognition. In their study, that focuses on both iris and retinal biometrics, they address the issue of lack of proper databases by taking advantage of the UBIRIS dataset, that incorporates 200 by 150 pixel infrared images of 241 irides. Out of this database, a subset of 17 irides was chosen and digital image processing was applied to make photographs resemble certain eye illnesses. As for the iris-related part of the study, following modifications were performed: keratitis and infiltrates, blurring and dulling of the cornea, scarring and surgery of the cornea, angiogenesis, tumors and melanoma. Authors report a very high FNMR values (32.8 -- 86.8\%) for all proposed modifications, except for the pathological vascularization (FNMR=6.6\%), changes in iris color (FNMR=0.5\%) and iridectomy-induced scarring in the iris (FNMR=0\%). They point to the segmentation problems (especially when cornea clouding is present) as a main cause of performance degradation. This study, however, does not take into account the fact that near-infrared light can penetrate the occluded cornea to a greater degree when compared to visible illumination and therefore enable correct imaging of the underlying iris structures \cite{Aslam}.

When looking at the current situation in scientific community around iris recognition one may assert that the issue of ocular disease influence on the performance of iris recognition systems is far from being fully investigated. There is little research on the topic, in most cases incorporating a small number of subjects (14 to 69 eyes) and only one iris coding methodology. Furthermore, only one research includes dataset of images representing more that one illness with real, not digitally synthesized data. Future studies should therefore focus on creating as vast databases as possible, including large number of ocular conditions in many ophthalmology patients. This should go side by side with incorporating more than one iris recognition methodologies to cross-evaluate their performance in presence of selected eye diseases.

\section{Database}

\subsection{Rationale behind creation of a dataset}
For the purpose of this study we created our own database, consisting of NIR-illuminated images of irides and in selected cases also photographs taken in visible light. This is to assert whether the NIR light allows better imaging of the underlying iris features than the visible light, in cases when certain occlusion, such as corneal haze or opacified lens is present. In addition to this, we gathered all the information that is typically associated with clinical practice, that being medical history and case description.

\subsection{Construction of the system used for data collection}
To make data collection as easy as possible we built a prototype imaging system that integrates an infrared iris recognition camera, a digital single-lens reflex camera (dSLR) and a laptop. The prototype was equipped with software to minimize the effort when conducting an ophthalmological examination by providing seamless and easy experience. 

\begin{figure}
\centering
\begin{subfigure}{0.5\textwidth}
  \centering
  \includegraphics[width=0.99\linewidth]{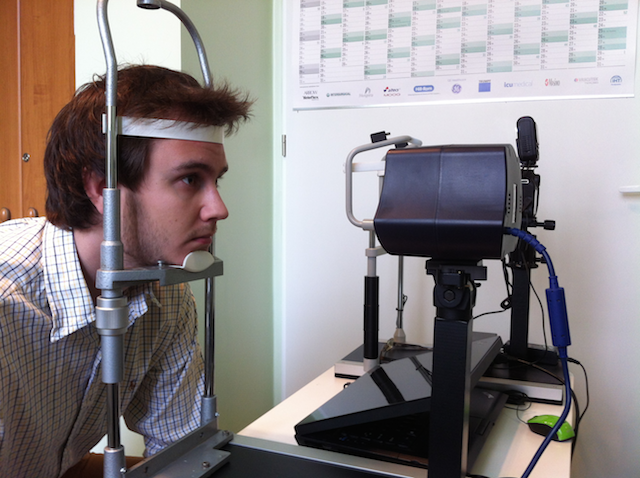}
  \caption{Subject having his eyes imaged in NIR light.}
  \label{nir:subject}
\end{subfigure}%
\begin{subfigure}{0.5\textwidth}
  \centering
  \includegraphics[width=0.99\linewidth]{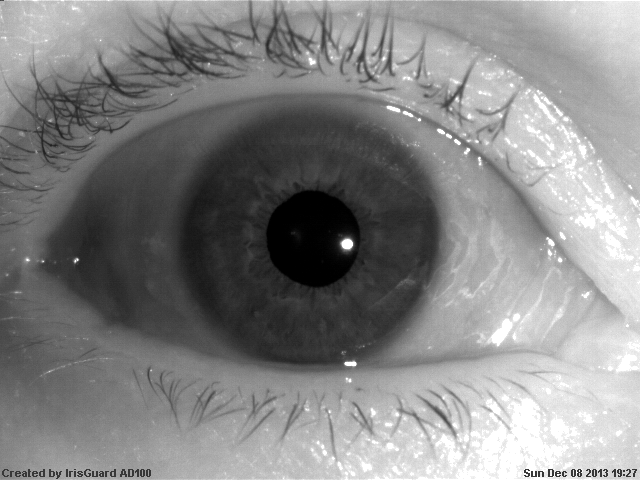}
  \caption{Sample NIR-illuminated iris image.}
  \label{nir:eye}
\end{subfigure}
\vskip0.2cm
\caption{NIR light imaging produces high quality iris images, compliant to ISO/IEC 19794-6 quality requirements.}
\label{nir}
\end{figure}

For NIR imaging, we use a commercially available IrisGuard AD-100 iris recognition camera attached to a stand equipped with chin and forehead rests. This is to keep a fixed distance between the eye and the camera to speed up the automatic localization of the iris incorporated in the AD-100 device and thus also the image capturing process. The complete setup is shown in Fig. \ref{nir:subject} and a sample NIR photograph is shown in Fig. \ref{nir:eye}.

\begin{figure}
\centering
\begin{subfigure}{0.5\textwidth}
  \centering
  \includegraphics[width=0.99\linewidth]{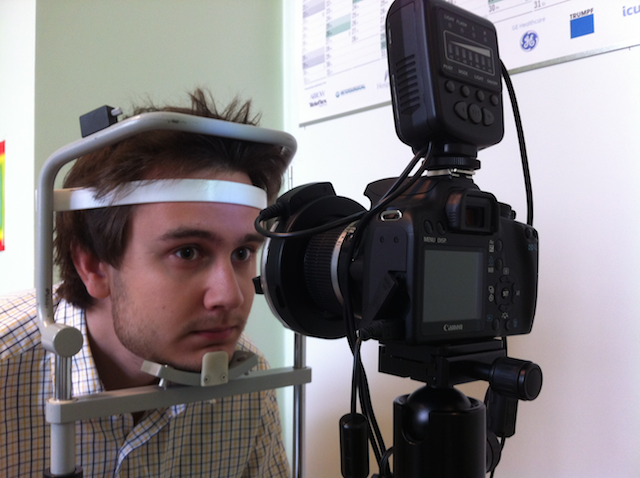}
  \caption{Subject having his eyes photographed.}
  \label{visible:subject}
\end{subfigure}%
\begin{subfigure}{0.5\textwidth}
  \centering
  \includegraphics[width=0.99\linewidth]{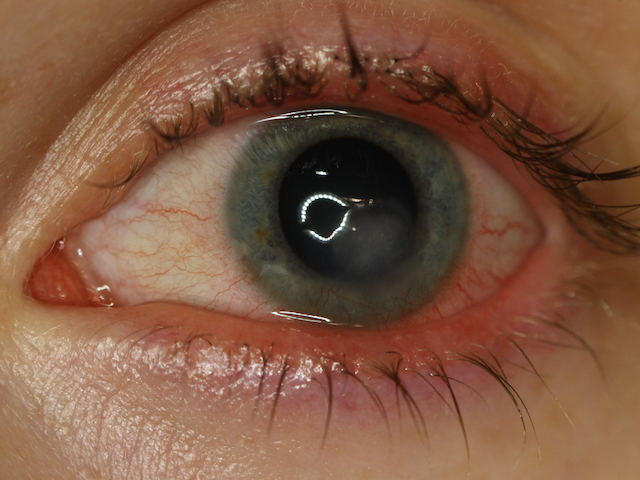}
  \caption{Sample eye image captured in visible light.}
  \label{visible:eye}
\end{subfigure}
\vskip0.2cm
\caption{Visible light imaging. Corneal clouding obstructing the view of iris is visible in the sample image to the right.}
\label{visible}
\end{figure}

As for the visible light illuminated images, we use a Canon EOS1000D dSLR camera equipped with a Canon EF-S 18-55mm f/4.5-5.6 IS lens, a macro lens converter to enable capturing a precise, magnified image of the eye and a ring LED flashlight designed for macro-photography. Also, a modeling light was built into the flash casing to enable automatic focusing in low-light conditions that are common in ophthalmology offices. This setup was placed on a stand same as the NIR camera. As the Canon camera is operated manually by the ophthalmologist it was equipped with a ball-head mounting to easily switch from one eye to another when capturing images. See Fig. \ref{visible:subject} for the final setup and Fig. \ref{visible:eye} for sample eye image.

These two devices are hooked up to a laptop running Windows$^\textrm{\tiny\textregistered}$ and our custom software designed for effortless gathering and managing patient and visit data. Both cameras are triggered from within the software, which eliminates the need to use multiple applications and therefore makes the examination more time effective. Acquired images are then being automatically downloaded to the computer, given timestamps and disease ID and stored on a hard drive. A preview is presented to the operator to evaluate the quality of an image and retry if necessary. Local MySQL instance is used to store patient and visit data and connect them to corresponding images using a relation database.

For selected cases we also take advantage of images collected using a Topcon DC3 slit lamp camera. This is used widely in an ophthalmological practice and allows better magnification of particularly interesting regions in the eye and thus reveals more details. Figures \ref{cataract} and \ref{cataract_accompanying} represent images acquired using this system.

\subsection{Data collection process}
Images were collected during the routine ophthalmology visits of patients of the Department of Ophthalmology of the Medical University of Warsaw, for both cataract patients and people suffering from other eye pathologies. All patients have been provided with information about this study and a written consent has been obtained from each person.

Apart from the typical ophthalmology examination, an eye image acquisition was performed. During subject's first visit he or she was registered into the system and automatically given a unique ID number. After a visit was created, at least six NIR-illuminated images and six visible light-illuminated images of each eye were captured. Each image was taken in a separate attempt, \emph{i.e.} the patient had to lift his head from the chin rest after each image capture, to intentionally introduce noise in intra-session sample sets. This course of action was then reiterated during all future visits.

\subsection{Database summary}
The resulting dataset consists of 1288 images, both NIR (480 $\times$ 640 pixel bitmaps) and visible light illuminated (10 megapixel, JPG-compressed), acquired from 37 patients. These images represent 11 different ocular pathologies. Out of this dataset we choose a subset of images acquired from patients suffering from cataract, later on referred to as the \emph{cataract group}.

\subsection{Data censoring}
Data in the cataract group was carefully evaluated and screened for poor quality images. We made sure that all images involved in the experiments are compliant to the ISO/IEC 19794-6:2011(E) and ISO/IEC 29794-6:201x(E) geometrical quality requirements \cite{ISO, ISO2}, so that images of insufficient quality do not cloud experimental results and only eye pathology related changes are taken into account. In particular, images with  following flaws were excluded from the final dataset:
\begin{itemize}
\item images that do not show the iris,
\item images with eyelashes or eyelids obstructing the iris (when there was less than 70\% of the iris visible),
\item images with poor focus or motion blur present,
\item images presenting pupil-to-iris diameter ratio outside the range of 0.2 to 0.7 (this is particularly important due to the use of mydriatics in clinical treatment).
\end{itemize}

After censoring, the resulting dataset of images in the cataract group consisted of 209 NIR illuminated images of irides collected from 35 eyes with illness present and 59 NIR-illuminated images of irides collected from 9 healthy eyes that we use as a control group. Those two subsets are later referred to as the \emph{disease group} and the \emph{control group}. Disease and control groups samples may partially originate from the same subjects (yet certainly not the same eyes).

\section{Iris recognition methods}
In our experiments we use three different, commercially available iris recognition methodologies, namely the BiomIrisSDK developed by Czajka \cite{BiomIrisSDK, CzajkaPacut}, the Neurotechnology's VeriEye SDK \cite{VeriEye} and MIRLIN SDK developed by Smart Sensors \cite{MIRLIN}.

\subsection{BiomIrisSDK}
The first matcher involved in this study incorporates automatic iris localization based on a modified Hough transform to find the boundary between the pupil and the iris and a modified Daugman's integro-differential operator for locating the boundary between the iris and the sclera. This methodology utilizes Zak-Gabor wavelet packets and derives binary iris feature vectors from the one-bit coding of the Zak-Gabor coefficients' signs. Hamming distance is employed to calculate the similarity score between two samples, spanning from 0 (when iris codes are identical) to 1 (when they are opposite to one another). HD value near zero denotes a good match, and values around 0.5 are typically obtained for different irides.

Apart from calculating the similarity scores between images, for the BiomIrisSDK we incorporated its low-level functions to ensure correctness of the segmentation process by generating polar images based on manually adjusted iris localization. This is to evaluate whether any potential impact that cataract may have on iris recognition is derived from segmentation issues, caused for example by too bright pupil regions in the image. We also manually created binary masks (\emph{i.e.} images denoting the occluded and unoccluded regions of the polar iris image) for each polar image to guarantee that only the portions of image that represent the actual iris pattern are taken into account during encoding. Those two subsets are later referred to as the \emph{automatic segmentation subset} and the \emph{manual segmentation subset}.

\subsection{VeriEye SDK}
Neurotechnology's VeriEye SDK makes use of an unpublished coding methodology and an off-axis iris localization incorporating active shape modeling. This method yields a similarity score for the two samples, ranging from 0 (non-match) to infinity (perfect match). However, in our experiments we observe 1~557 as a maximum value when comparing two identical images.

\subsection{MIRLIN SDK}
Lastly, we use Smart Sensors' MIRLIN SDK, which is a commercially available method that utilizes differences of discrete cosine transform coefficients calculated for overlapped angular patches of normalized iris images \cite{DCT}. As for the BiomIrisSDK matcher, this methodology also employs Hamming distance as a dissimilarity metric.

\section{Results}

\subsection{Calculation of similarity scores}
For each of the available matchers we created a distribution of similarity scores calculated from all possible pairs of same-eye images (\emph{i.e.} genuine comparisons), as well as from all possible pairs of images showing different eyes (\emph{i.e.} impostor comparisons). This was performed for both control and disease group of samples. For the BiomIrisSDK matcher we also generated an additional set of comparisons using the dataset with manually corrected segmentation. When generating comparison scores (both genuine and impostor) we used each image pair only once (if sample A was matched to sample B, then the opposite matching was not performed, \emph{i.e.} sample B was not matched to sample A).

For the BiomIrisSDK and VeriEye SDK matchers it was possible to produce 177 and 601 genuine comparison scores in the control group and disease group, respectively. For those matchers we also generated 1~534 and 21~344 impostor scores in the control group and disease group, respectively. For the MIRLIN matcher these numbers are slightly lower due to the template extraction errors in several cases (therefore such samples were excluded from the MIRLIN-related results): the MIRLIN matcher generated 177 and 585 genuine scores (control and disease group, respectively) as well as 1~534 and 20~943 impostor scores (control and disease group, respectively).

In the next subsection we present the average similarity scores of the calculated genuine and impostor score distributions for the disease group when compared to the control group of healthy eyes. The cumulative distribution functions visualize a possible disease-related shift in these distributions that would indicate if the cataract-induced changes in the eye can indeed contribute to the degradation of iris recognition system performance. For the BiomIrisSDK matcher we also present a comparison of scores obtained using automatic versus manual iris image segmentation to assess if faulty segmentation is a contributing factor.

\subsection{Score distribution comparison}

\subsubsection{VeriEye SDK}
\begin{figure}[!h]
\centering
\begin{subfigure}{0.5\textwidth}
  \centering
  \includegraphics[width=0.99\linewidth]{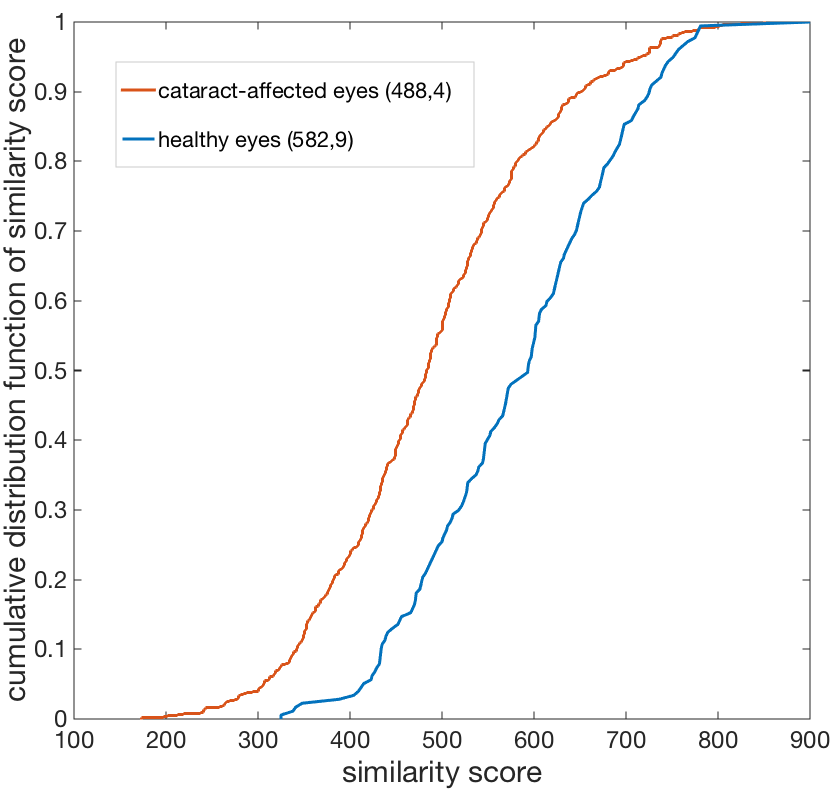}
  \caption{Genuine distributions.}
  \label{ve_genuines}
\end{subfigure}%
\begin{subfigure}{0.5\textwidth}
  \centering
  \includegraphics[width=0.99\linewidth]{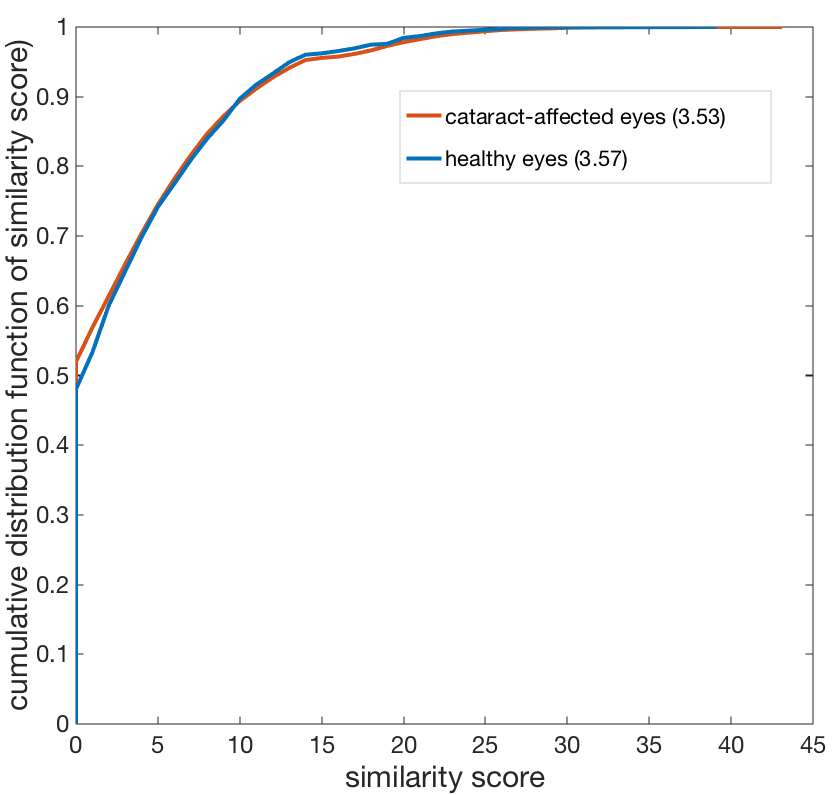}
  \caption{Impostor distributions.}
  \label{ve_impostors}
\end{subfigure}
\vskip0.2cm
\caption{Comparison of genuine (a) and impostor (b) score distributions in in control group versus disease-affected group for the VeriEye matcher. Higher score denotes a better match. Average similarity scores for each distribution are denoted in brackets.}
\end{figure}

Using the VeriEye SDK matcher we observe a decrease in average genuine similarity score by approximately 16.2\% when disease and control groups are compared. Figure \ref{ve_genuines} shows a visible shift in score distributions of the cataract-affected eyes towards worse similarity scores (\emph{i.e.} towards left side of the graph, as lower score denotes a worse match). This observation is statistically significant in terms of the unbalanced one-way analysis of variance (namely the ANOVA test) that casts doubt on a null hypothesis that the obtained scores are derived from distributions with the same mean values ($p$-value~$<10^{-19}$). We do not observe statistically significant change in the impostor distributions (average score of 3.57 in the control group compared to the 3.53 in the disease group), Fig. \ref{ve_impostors}. This is confirmed by the ANOVA test, since we have no rationale behind rejecting the null hypothesis on the mean equality ($p$-value~$=0.7687$).

\subsubsection{MIRLIN SDK}
\begin{figure}[!h]
\centering
\begin{subfigure}{0.5\textwidth}
  \centering
  \includegraphics[width=0.99\linewidth]{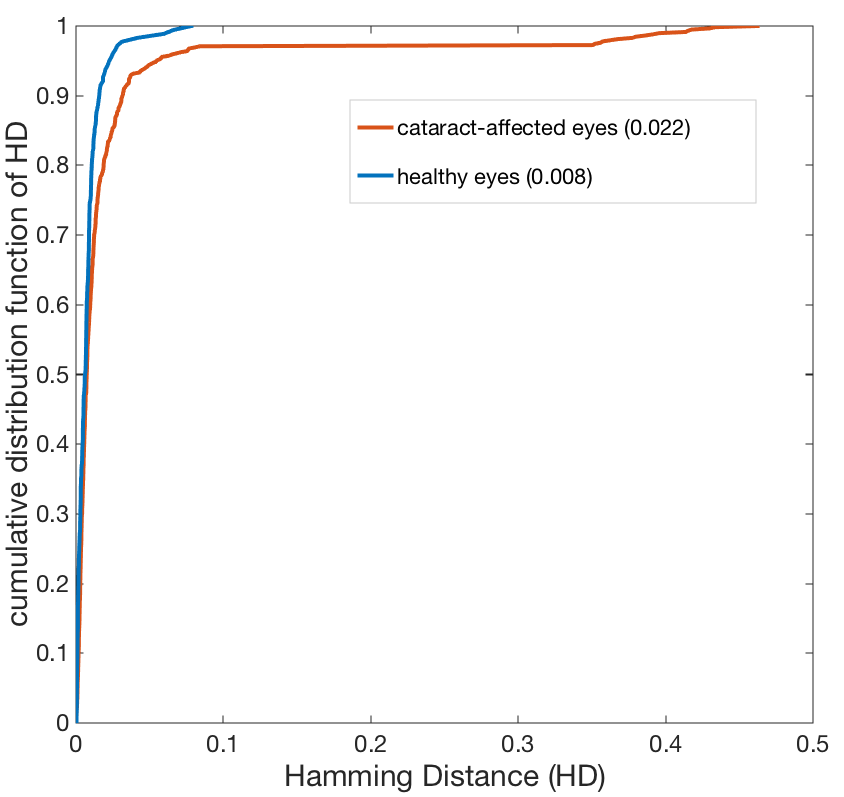}
  \caption{Genuine distributions.}
  \label{mir_genuines}
\end{subfigure}%
\begin{subfigure}{0.5\textwidth}
  \centering
  \includegraphics[width=0.99\linewidth]{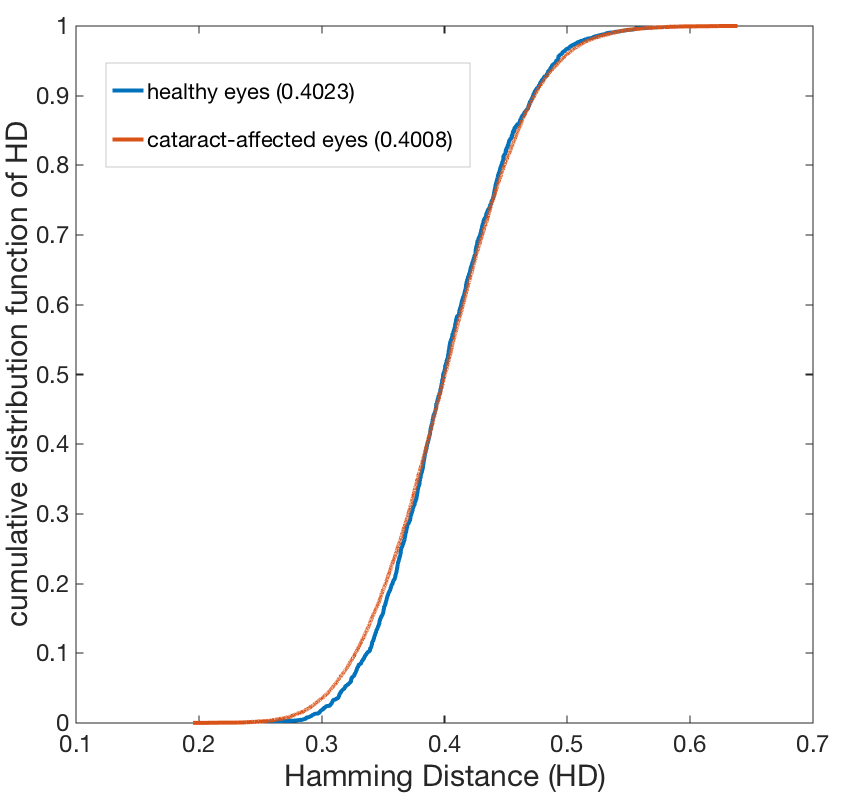}
  \caption{Impostor distributions.}
  \label{mir_impostors}
\end{subfigure}
\vskip0.2cm
\caption{Comparison of genuine (a) and impostor (b) scores distributions in in control group versus disease-affected group for the MIRLIN matcher. Lower score denotes a better match. Average similarity scores for each distribution are denoted in brackets.}
\end{figure}

As for the MIRLIN matcher, we show a statistically significant deterioration of the average genuine score by over 175\% for the disease group when compared to the control group (corresponding variance analysis leads to $p$-value~$=0.0038$). See Figure \ref{mir_genuines} for a visible shift between these two distributions with the disease group distribution shifted to the right, that is towards worse (\emph{i.e.} larger) Hamming distance scores. Distributions of impostor comparisons did not reveal any statistically significant changes (mean value of 0.4008 in the disease group compared to 0.4023 in the control group, and ANOVA's $p$-value~$=0.3119$, Fig. \ref{mir_impostors}).

\subsubsection{BiomIrisSDK}
\begin{figure}[!h]
\centering
\begin{subfigure}{0.5\textwidth}
  \centering
  \includegraphics[width=0.99\linewidth]{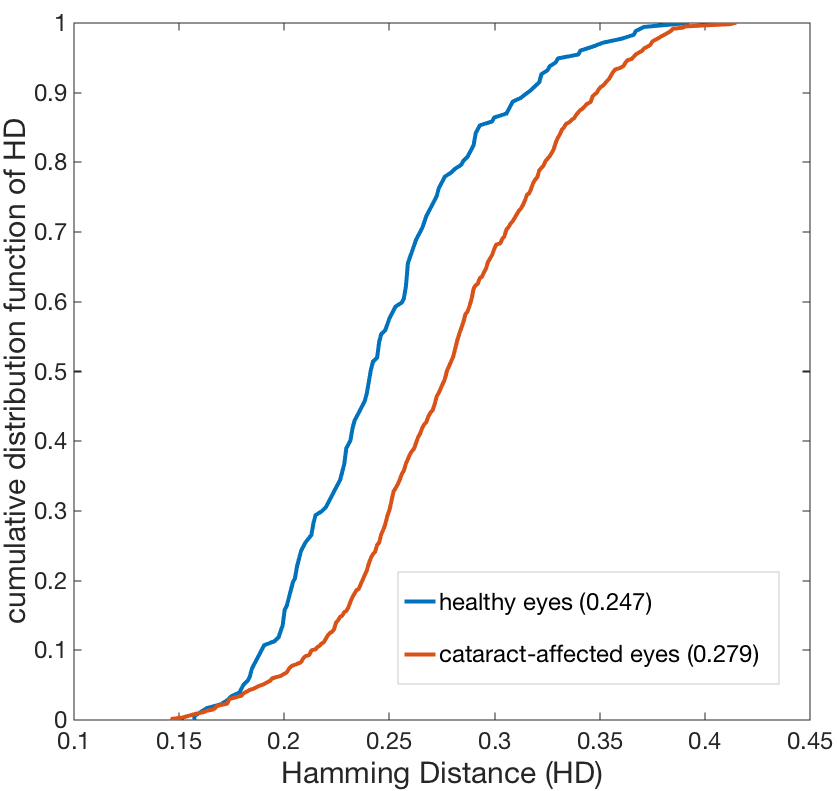}
  \caption{Genuine distributions.}
  \label{ac_genuines_auto}
\end{subfigure}%
\begin{subfigure}{0.5\textwidth}
  \centering
  \includegraphics[width=0.99\linewidth]{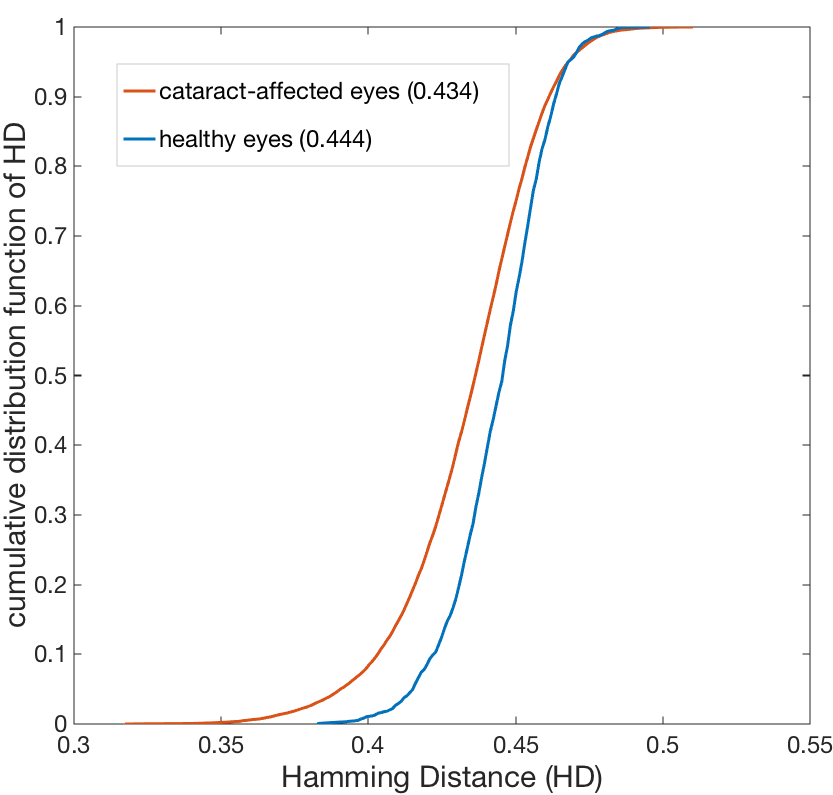}
  \caption{Impostor distributions.}
  \label{ac_impostors_auto}
\end{subfigure}
\vskip0.2cm
\caption{Comparison of genuine (a) and impostor (b) scores distributions in in control group versus disease-affected group for the BiomIrisSDK matcher. Automatic segmentation subset is used. Lower score denotes a better match.}
\label{ac_auto}
\end{figure}

Using the BiomIrisSDK matcher with automatic segmentation leads to observation of degradation in average Hamming distances by about 12.6\% when comparing scores of disease group to scores obtained from control group (as shown in Figure \ref{ac_genuines_auto}). The ANOVA test questions the null hypothesis that investigated sample subsets are drawn from populations with the same mean values (since the $p$-value~$<10^{-11}$). For the impostor distribution we observe a degradation of about 2.3\%, and in contrast to the former matchers this change is statistically significant, since the ANOVA's $p$-value~$<10^{-60}$ (Fig. \ref{ac_impostors_auto}).

\begin{figure}[!h]
\centering
\begin{subfigure}{0.5\textwidth}
  \centering
  \includegraphics[width=0.99\linewidth]{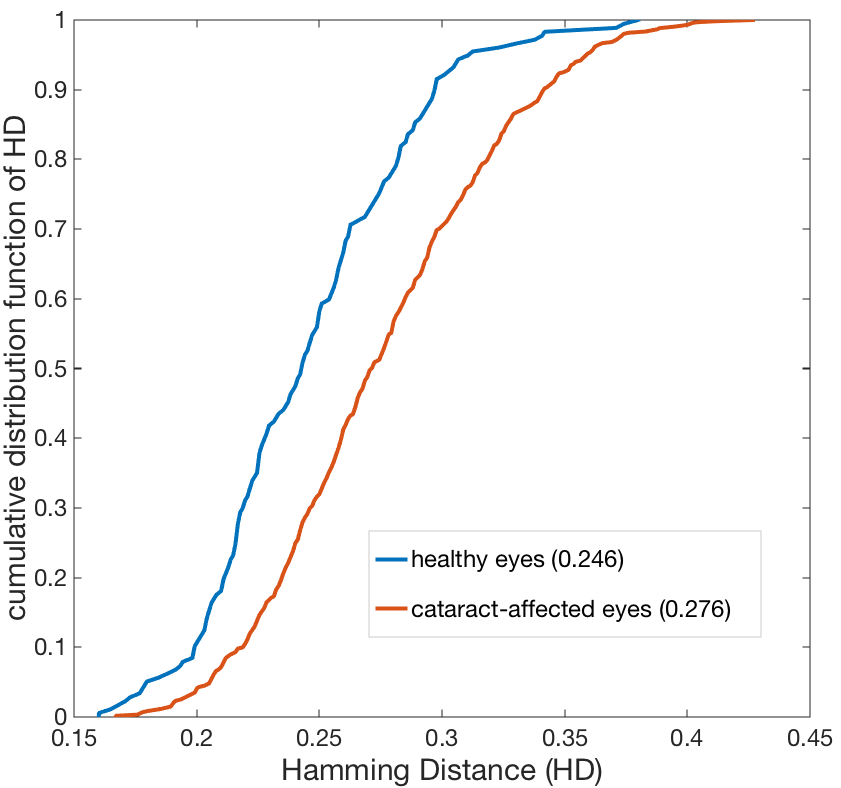}
  \caption{Genuine distributions.}
  \label{ac_genuines_manual}
\end{subfigure}%
\begin{subfigure}{0.5\textwidth}
  \centering
  \includegraphics[width=0.99\linewidth]{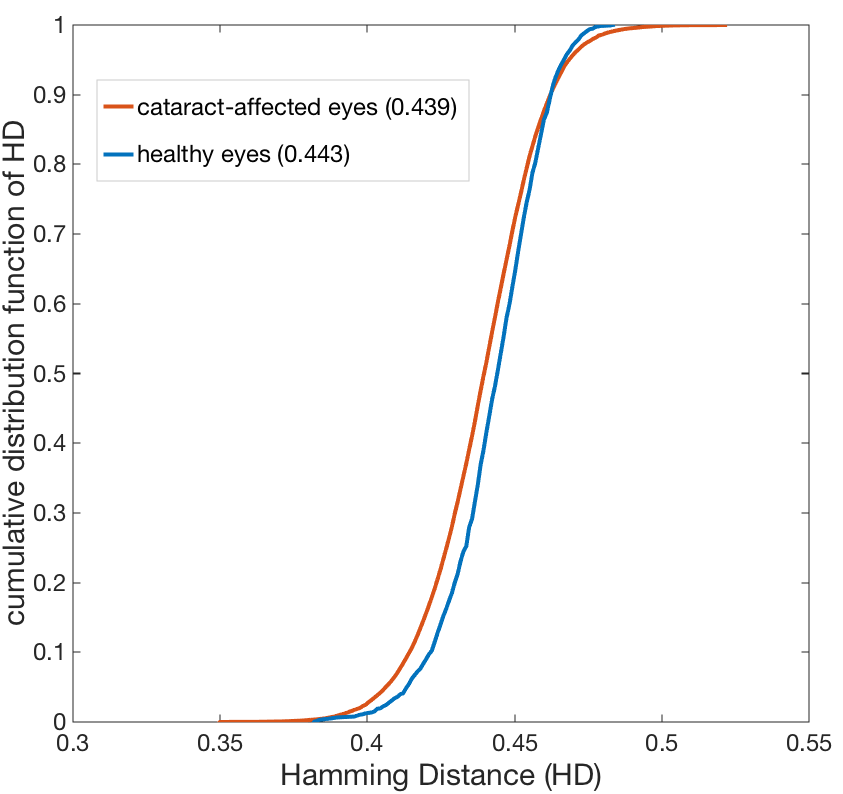}
  \caption{Impostor distributions.}
  \label{ac_impostors_manual}
\end{subfigure}
\vskip0.2cm
\caption{Same as in Fig. \ref{ac_auto}, but manual segmentation subset is used.}
\end{figure}

When evaluating the results obtained using the manual segmentation subset we find that the average genuine score decreased by 12.3\%, as confirmed by the ANOVA test with $p$-value~$<10^{-13}$ (see Figure \ref{ac_genuines_manual}). The impostor distribution using this subset showed an increase in average Hamming distance score of about 0.9\% (Fig. \ref{ac_impostors_manual}). Based on the outcome of the ANOVA test, also in this case we cannot accept the null hypothesis that samples are drawn from distributions with the same mean value ($p$-value~$<10^{-13}$).

\begin{figure}[!h]
\centering
\begin{subfigure}{0.5\textwidth}
  \centering
  \includegraphics[width=0.99\linewidth]{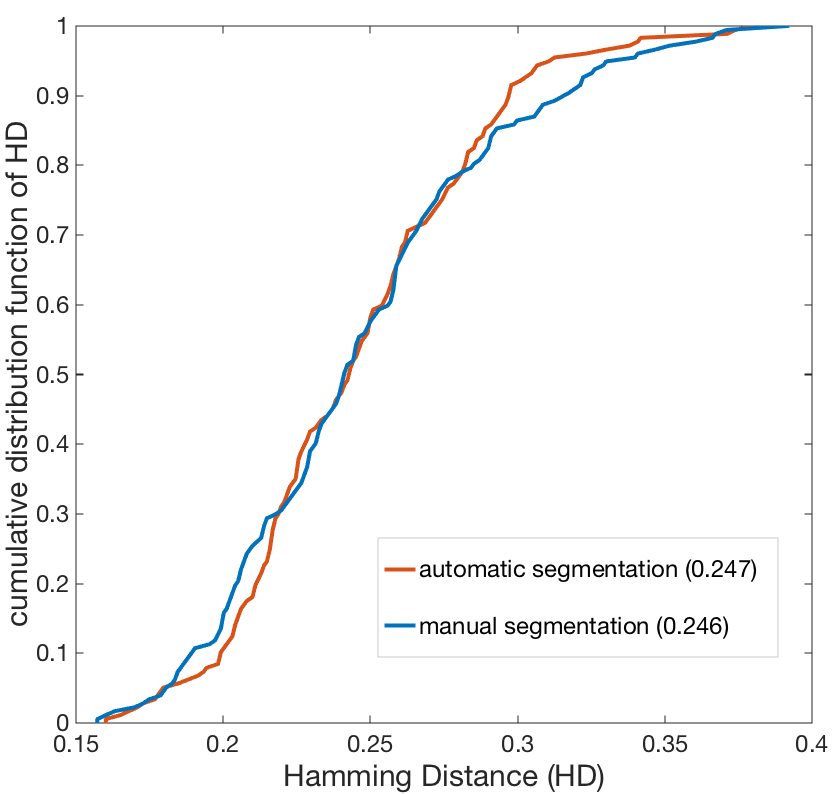}
  \caption{Control group.}
  \label{ac_control_segmentation}
\end{subfigure}%
\begin{subfigure}{0.5\textwidth}
  \centering
  \includegraphics[width=0.99\linewidth]{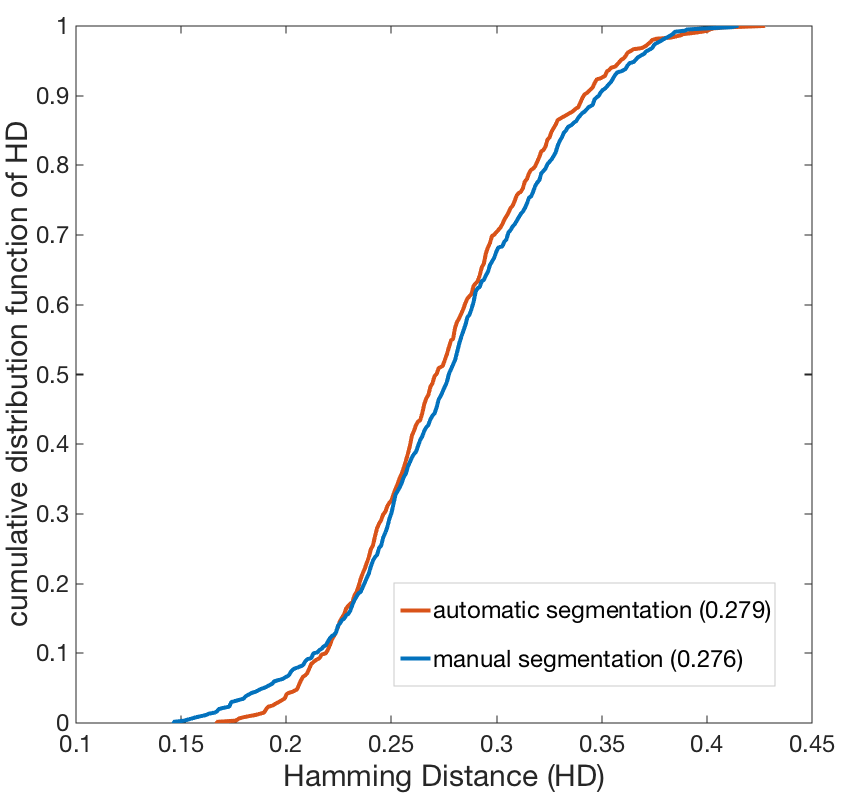}
  \caption{Disease group.}
  \label{ac_disease_segmentation}
\end{subfigure}
\vskip0.2cm
\caption{Comparison of genuine scores obtained using automatic and manual segmentation subsets of control (a) and disease (b) group for the BiomIrisSDK matcher. Mean values for each distribution are denoted in brackets.}
\end{figure}

Lastly, we compare the genuine distributions of results using manual and automatic segmentation subsets for both control and disease group (see Figures \ref{ac_control_segmentation} and \ref{ac_disease_segmentation}, respectively). In both cases there is little difference between scores derived from automatic and manual segmentation subsets and based on the outcome of the ANOVA test we cannot discard the null hypotheses that samples are drawn from populations with the same mean values ($p$-value~$=0.796$ and $p$-value~$=0.439$ for control and disease group, respectively). This suggests that applying manual correction of the segmentation results has little effect on the matcher performance, and the degradation may come from the illness itself.

\subsection{Discussion}
Results obtained from our experiments show visible and statistically significant shifts towards worse similarity scores in genuine distributions obtained from the disease group when compared to those derived from the control group. These observations are true for each of the tested methods and have been confirmed using the one-way unbalanced analysis of variance with \emph{p}-values below the significance threshold $\alpha=0.05$ in each case. This clearly shows that eyes suffering from cataract may cause a degradation in iris recognition performance, independently of the employed recognition methodology.

For the VeriEye matcher this does not translate into degrading the performance in terms of false rejected or false accepted samples. For the MIRLIN matcher 3\% of the samples in the disease group produce scores beyond 0.32, that is the typical acceptance threshold for methods incorporating Hamming distance as a similarity metric. Those samples would thus be false rejected if the acceptance threshold is set to 0.32. There are no false accepted samples for this methodology. For the BiomIrisSDK matcher we find an increase in FNMR in disease group when compared to the control group from 9.6\% to 22.5\% when automatic segmentation is employed, and from 4.5\% to 19.5\% with the use of manual segmentation. Again, we observe no false acceptances in either of the subsets. This may suggest that selected matchers (in our study, this is true for the Neurotechnology's matcher) are somehow resilient to changes induced by this eye pathology and it may not necessarily cause them to degrade performance in terms of false acceptance and false rejection of samples, despite of the underlying degradation in similarity scores. Some matchers, however, do produce worse results affecting an overall system's performance.

When evaluating the effects of applying manual segmentation correction we find that it has little influence on the degradation of matcher's performance (with differences between the two distributions not showing statistical significance in terms of the ANOVA test). This may suggest that there are other factors that affect recognition reliability apart from the faulty segmentation. Those may include a deviation from pupil's circularity, iris pattern obstructed by pathological changes in the eye, or some alterations to the iris features from which iris templates are derived, that may not be easy to perceive by a human observer in visible light.

To suggest particular conditions that may have a strong effect on iris recognition reliability we also performed a detailed visual observation in cases with significantly bad recognition performance. To find those `low quality' samples, we consider images in the disease group generating genuine scores that are worse than the worst genuine score obtained in the control group. For the VeriEye SDK 9 eyes produced comparison scores that met this criteria (\emph{i.e.} the similarity score below 325). Out of those, 3 eyes had pupil dilated more than the rest, but with pupil-to-iris diameter ratio still falling between values 0.2 and 0.7 as specified by the ISO/IEC \cite{ISO2}. For the next two eyes, only one image of each of them produced poor scores. In the first case, the eye was obstructed by hair and in the latter it was gazing away (optical axis of the eye was not aligned with the optical axis of the camera). However, four remaining eyes did not show any visible change to the iris pattern, deviation from pupil's circularity, excessive dilation/constriction of the pupil or any other flaws that might indicate the source of poor matcher's performance.

As for the MIRLIN matcher, 4 eyes in the disease group produced scores higher (as higher score denotes a worse match) than the highest score obtained from the control group, that is above 0.0782. One of them is the same gazing away eye that affected the VeriEye matcher. One eye had an upper eyelid located very low, but the amount of visible iris tissue was still exceeding 70\%, as suggested by the ISO/IEC standard. The remaining two eyes did not have any visible changes.

When the BiomIrisSDK matcher is involved, we distinguish the automatic and manual segmentation subsets. In the disease group of the former, scores obtained from 3 eyes were higher than the highest score in the control group (\emph{i.e.} 0.3916). Two of those eyes did not have any visible changes. The third eye had a pupil that was not exactly circular and lens clouding was present. This eye did not, however, produce high scores when manual segmentation has been applied. This may point to the hypothesis that the illness-related deviation from pupil circularity affected the segmentation process and therefore this eye produced higher scores leading to false rejection. In the disease group of the manual segmentation subset 5 eyes gave scores above the highest score in the control group (that is above the value of 0.3799). One of them was the gazing away eye that showed high scores for both VeriEye and MIRLIN matchers. The remaining four eyes did not show any changes or alterations visible during visual inspection.

Photographs of eyes with excessive pupil dilation or constriction, with different gaze directions and objects obstructing the view of the iris will typically produce worse similarity scores than those perfectly compliant with the ISO/IEC image quality standards. For all tested methodologies, however, there are images of eyes yielding very poor scores, but at the same time not showing any apparent flaws, such as deviation form pupil's circularity or visible alterations to the iris pattern. This may suggest that cataract can cause effects that are not obvious to the human observer, but reveal themselves in terms of impact they have on iris recognition performance. In addition, for different matchers, different eyes represent such behavior (only 2 out of 10 eyes cause severe problems to more than one matcher), which may indicate that different methodologies are vulnerable to different, perhaps subtle changes to the appearance of the iris and its surrounding structures.

\section{Conclusion}
In this study we present experimental evidence supporting the hypothesis that cataract may have a negative influence on the reliability of iris recognition systems. Every matcher employed produced worse results when presented with images obtained from the disease-affected eyes and for two out of three used methodologies this also translated to an increase in false non-match rate. We also showed that this behavior can be unrelated to the faulty segmentation during image pre-processing, and therefore there are some other factors that contribute to the decrease in matcher's accuracy. 

These results, however, should not be a starting point to diminish iris recognition as a biometric application, or any method applied in this work as examples, as it simply refrains with impact caused by other illnesses to distinct biometric characteristics (like severe pharyngitis in voice recognition). This paper shows that one should be aware of possible negative influence of cataract on the biometric reliability, and the adequate countermeasures should be applied (like thorough observation of eyes being enrolled to the system). 

\acknowledgments        
This paper summarizes a part of Mateusz Trokielewicz's M.Sc. project realized at the Warsaw University of Technology in years 2013--2014, and lead by Dr. Adam Czajka. Authors would like to thank Ms. Weronika Gutfeter, Mr. Krzysztof Kozio\l \, and Mr. Krzysztof Piech for their help and insight when building the hardware and software used for database collection.

\end{document}